\newcommand{\CLASSINPUTtoptextmargin}{0.75in}
\newcommand{\CLASSINPUTbottomtextmargin}{1.01in}

\documentclass[conference]{IEEEtran}
\IEEEoverridecommandlockouts

\usepackage{cite}
\usepackage{amsmath,amssymb,amsfonts}
\usepackage{algorithmic}
\newtheorem{definition}{Definition}
\usepackage{graphicx}
\usepackage{textcomp}
\usepackage{xcolor}
\usepackage{tcolorbox}
\usepackage{enumitem}
\setlist[itemize]{left=0pt}
\setlist[enumerate]{left=0pt}
\usepackage{xspace}

\usepackage{multirow}
\usepackage{booktabs}
\definecolor{tabletitleloss_color}{HTML}{EF6C63}
\usepackage{colortbl}
\usepackage{placeins}

\definecolor{table_result_best}{HTML}{FEECE7}
\definecolor{table_result_2nd}{HTML}{DEEEED}
\newcommand{\colorblockTab}[2][0.5cm]{\textcolor{#2}{\rule{#1}{#1}}}

\definecolor{tab_seen}{HTML}{875C92}

\newcommand{\FML}{\ensuremath{\xspace\texttt{\$FML}}\xspace}

\def\BibTeX{{\rm B\kern-.05em{\sc i\kern-.025em b}\kern-.08em
    T\kern-.1667em\lower.7ex\hbox{E}\kern-.125emX}}
\begin{document}

\title{Multi-Continental Healthcare Modelling Using Blockchain-Enabled Federated Learning\\
\thanks{This research is supported in part by grants from the National Science Foundation of China (No. 42171404), and Shanghai Engineering Research Center for Blockchain Applications And Services (No. 19DZ2255100).}
}

\author{\IEEEauthorblockN{Rui Sun$^{1}$, Zhipeng Wang$^{2}$, Hengrui Zhang$^{3}$, Ming Jiang$^{4}$, Yizhe Wen$^{5}$, \\Jiahao Sun$^{5}$, Erwu Liu$^{*}$$^{4,6}$, and Kezhi Li$^{*}$$^{3,7}$}
\thanks{$^{*}$corresponding author e-mail: xinyuqu@tongji.edu.cn, ken.li@ucl.ac.uk}
\\
\IEEEauthorblockA{$^{1}$ School of Computing, Newcastle University, Newcastle upon Tyne, United Kingdom}  
\IEEEauthorblockA{$^{2}$ Department of Computing, Imperial College London, London, United Kingdom}
\IEEEauthorblockA{$^{3}$Institute of Health Informatics, University College London, London, United Kingdom}  
\IEEEauthorblockA{$^{4}$Shanghai Research Institute for Intelligent Autonomous Systems, Tongji University, Shanghai, China}
\IEEEauthorblockA{$^{5}$ FLock.io, London, United Kingdom} 
\IEEEauthorblockA{$^{6}$College of Electronics and Information
Engineering, Tongji University, Shanghai, China}  
\IEEEauthorblockA{$^{7}$University College London Hospital, London, United Kingdom}  
}

\maketitle

\begin{abstract}
One of the biggest challenges of building artificial intelligence (AI) model in the healthcare area is the data sharing. Since healthcare data is private, sensitive, and heterogeneous, collecting sufficient data for modelling is exhausting, costly, and sometimes impossible. In this paper, we propose a framework for global healthcare modelling using datasets from multi-continents (Europe, North America, and Asia) without sharing the local datasets, and choose glucose management as a study model to verify its effectiveness. Technically, blockchain-enabled federated learning is implemented with adaptation to meet the privacy and safety requirements of healthcare data, meanwhile, it rewards honest participation and penalizes malicious activities using its on-chain incentive mechanism. Experimental results show that the proposed framework is effective, efficient, and privacy-preserving. Its prediction accuracy consistently outperforms models trained on limited personal data and achieves comparable or even slightly better results than centralized training in certain scenarios, all while preserving data privacy. This work paves the way for international collaborations on healthcare projects, where additional data is crucial for reducing bias and providing benefits to humanity.
\end{abstract}


\section{Introduction}
While data-driven machine learning techniques have catalyzed breakthroughs in other domains, ~\cite{silver2017mastering,vaswani2017attention}, 
their progress in healthcare has been more incremental. This disparity is largely attributed to the unique challenges associated with healthcare data \cite{jiang2017artificial}.
Healthcare data is distinctively private, sensitive, and heterogeneous, presenting significant barriers not commonly encountered in other fields \cite{abouelmehdi2018big, narula2024comprehensive}. 
Accumulating sufficient data for effective training and testing of artificial intelligence (AI) models is not only challenging but also costly. The process of data collection is fraught with obstacles, varying significantly across global regions. For example, in the study of chronic diseases, researchers must establish partnerships with hospitals, private companies, or public authorities to gather sufficient data. Each partner, however, treats their data as a proprietary resource, complicating the sharing process. Moreover, the approval process for data access is often complex and protracted due to regulations and country barriers \cite{haque2021gdpr}. These factors collectively hinder the application of AI in healthcare and are increasingly becoming critical bottlenecks in the development of AI-enabled healthcare systems.
\begin{figure}
    \centering
    \includegraphics[width=1\linewidth]{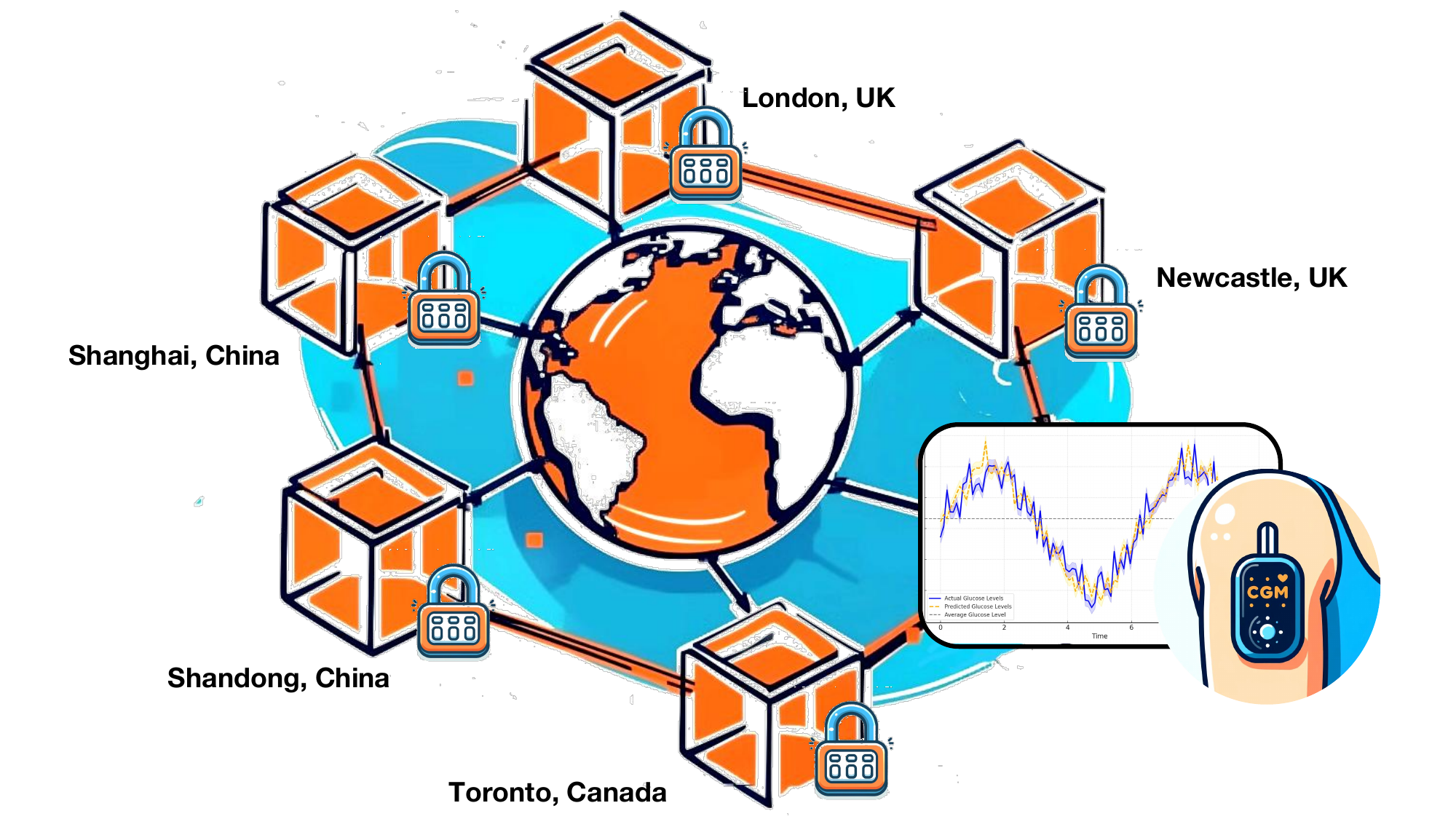}
    \caption{A diagram that illustrates blood glucose prediction modelling without sharing private data for five participants from three continents. }
    \label{fig:enter-label}
\end{figure}

To address the prevalent challenge of data privacy and access, we use a privacy-preserving framework for AI modeling that leverages federated learning and blockchain. We selected blood glucose prediction as a case study to illustrate the potential of this approach in healthcare applications. 
Approximately 422 million people globally suffer from diabetes, with 1.5 million deaths directly attributed to the disease annually \cite{lin2020global}. An effective strategy for mitigating severe complications due to hyperglycemia or hypoglycemia involves early intervention to regulate blood glucose levels for individuals with type 1 diabetes (T1D). Usually, the data obtained from clinical trials are typically scarce and costly to acquire \cite{glick2014economic}. Research indicates that universal models, trained on data pooled from groups of T1D patients, can significantly enhance glucose prediction accuracy \cite{piao2024garnn}. 

In this paper, we propose a Multi-Continental Glucose Prediction~(MCGP) framework using blockchain-enabled federated learning. This technique allows participating organizations to contribute to model training without the need to share sensitive health data directly. Instead, model parameters are exchanged among participants, thus preserving the privacy of the underlying health data. It also has protocols to apply incentive mechanisms to encourage good contributors and to build robustness in detecting malicious behaviors. For the first time, we tested the proposed modeling framework within a network spanning three continents—Europe, North America, and Asia—demonstrating that the model derived from this collaborative approach outperforms those trained on isolated datasets from individual health data, shown in Fig~\ref{fig:enter-label}. This framework not only proves effective for blood glucose prediction but can also be readily extended to model other chronic diseases or address broader healthcare challenges.


\section{Problem Formulation and Existing Works}


There are three components in the work, including blood glucose level prediction (BGLP), decentralized federated learning, and blockchain deployment. Here we provide definitions for each of the components. 

\subsection{BGLP}
The goal of BGLP is to predict future glucose levels using current and historical data, including glucose levels (mg/dL) collected from continuous glucose monitors (CGMs), and optional recorded data such as meal intake (grams as carbohydrate), insulin injection (mg or units) and time of the day. Specifically, 
\begin{definition}
$\bold {BGLP}$: Given a time series $X$ of historical glucose records $g_{1:L} = \{ g_1, \dots, g_L\}$ and associated variables (such meal intake $m$, insulin $i$, and timestamp $t$) collected at regular time intervals $\tau$ (e.g. $\tau=5$ mins), BGLP aims to accurately predict future glucose level $g_{L+H}$ where $H$ denotes the predictive horizon (e.g. $H=6$ to predict the future glucose in $30$ mins), and $L$ can vary as a sliding window. 
\end{definition}

Given the substantial interpersonal and intrapersonal variability inherent in glucose data, training the BGLP model necessitates individual-specific datasets to develop tailored models that accurately reflect personal health metrics. 
However, the acquisition of individual data for machine learning applications can be both challenging and resource-intensive. 
An effective alternative involves the utilization of a population model, which leverages data from a broad demographic. 
Research indicates that population models, when trained with extensive datasets from numerous individuals, typically outperform personalized models, especially when personal data is scant and costly \cite{zhu2022personalized, piao2024garnn}. To fully capitalize on the advantages offered by the population model, it is proposed that a global collaborative framework be established whereby individuals, organizations, and healthcare institutions contribute their data to a collective model training effort in a manner that ensures the privacy of the contributors' data. 

\subsection{BGLP Using Federated Learning (FL)}
 In this context, federated learning becomes a useful approach. It enables the decentralized training of machine learning models wherein each participant (e.g., a hospital or an individual) retains control over their own data. Only model updates, such as gradients or learned parameters, are shared centrally for aggregation, without transmitting any personal or sensitive data. 
FL promises a solution that enables numerous participants (e.g., hospitals and individual patients) to collaboratively train a global ML model while preserving local data privacy \cite{rieke2020future}. During FL training of BGLP, only participants' local model updates, such as gradients or learned parameters of BGLP models \cite{de2023federated}, are shared for knowledge aggregation without transmitting any personal or sensitive local data. 

\begin{definition}
\textbf{BGLP Using FL}: Consider a federated system with $K > 1$ participants(hospitals in this paper), denoted by the set $\mathcal{K} = {1, 2, \ldots, K}$. Let $\mathcal{D}_k$ represent the local data stored by participant $k$, where $\mathcal{D}_k \cap \mathcal{D}_l = \varnothing$ for $k \neq l$ and $k, l \in \mathcal{K}$. Each local dataset $\mathcal{D}_k$ can be randomly split into a training set $\mathcal{D}^{\text{train}}_k$, a validation set $\mathcal{D}^{\text{val}}_k$, and a test set $\mathcal{D}^{\text{test}}_k$, all of which remain private to participant $k$. The global model is denoted as $\theta_g$, and each participant’s local model is denoted as $\theta_k$. The total number of global communication rounds is $R$, with each round consisting of a single global aggregation. The number of local model update epochs per communication round is denoted by $E$.
\end{definition}



\subsection{Blockchain-Enabled FL}
The properties of decentralization and transparency of blockchain technology can further improve the decentralization and security of FL.
In the following, we provide the definition of blockchain-enabled FL.
\begin{definition}
\textbf{Blockchain-enabled FL}: Blockchain-enabled FL (BCFL) represents an innovative paradigm of blockchain technology with FL, aiming to enhance security and trust in decentralized machine learning environments. In this paradigm, blockchain serves as an immutable ledger, recording transactions and models exchanged across the distributed nodes participating in the federated learning process. This integration addresses core challenges such as data privacy, security, and model integrity, by ensuring transparent and verifiable transactions while maintaining the confidentiality of the data. Blockchain's decentralized nature allows for a trustless system where no single entity has control over the entire dataset or the learning process, thereby mitigating risks associated with centralized data storage and processing. Moreover, the use of smart contracts automates the process of data sharing and model updates in FL. Additionally, blockchain's inherent incentive mechanisms reward honest participation and penalize malicious activities, further enhancing the security of FL systems. BCFL has garnered significant attention for its applicability in diverse fields, including the Internet of Things (IoT)~\cite{issa2023blockchain, kalapaaking2022blockchain} and healthcare~\cite{myrzashova2023blockchain, lakhan2022federated, dong2024defending}. 



\end{definition}

\section{Blockchain-Enabled Federated Learning Glucose Modelling}

\begin{figure*}[th]
\vspace{0.1cm}
  \centering
  \includegraphics[width=1\linewidth]{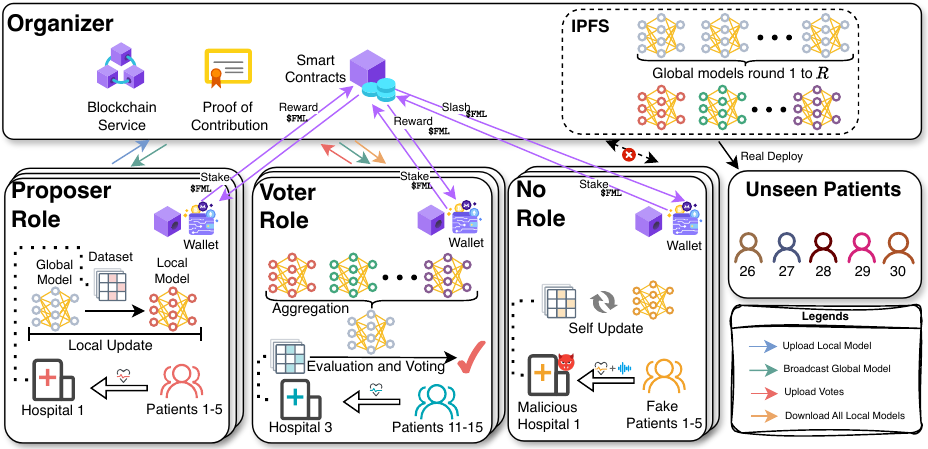}  
  \caption{Overview of the MCGP framework architecture. Participants (hospitals) are randomly assigned roles as either proposers or voters to collaboratively train a model using a reward and slashing mechanism. Malicious participants are identified during training, assigned no role, and removed from the FL training task. The final optimized global model is then deployed to test on unseen patients’ data.}
  \label{fig:overall_sys}
\end{figure*}

To model blockchain-enabled federated learning for glucose prediction, we designed the Multi-Continental Glucose Prediction (MCGP) Framework, shown in Fig~\ref{fig:overall_sys}. This framework leverages federated learning to enable different hospitals to collaboratively train a glucose prediction model while preserving patient data privacy. Blockchain technology is utilized to implement a reward and slashing mechanism, incentivizing honest participation and detecting malicious actors.

\subsection{Glucose Prediction}

Many approaches have been developed for glucose prediction. Here We formulate the BGLP problem as a multi-variant time series prediction problem as in \cite{zhu2022personalized}. A sliding window is utilized to crop historical data as input data. The future glucose value (e.g. in 30 mins) is used as the target output. 

\subsection{Federated Learning for Glucose Prediction}

In privacy-sensitive fields such as glucose management, FL can break down data silos among hospitals. 
Each patient has unique characteristics; for instance, insulin levels can vary significantly in response to eating, drinking water, and insulin injections. 
Each hospital’s prediction model needs to learn these personalized features to provide timely and accurate blood sugar predictions for new patients. FL enables hospitals to collaboratively enhance their models by sharing knowledge, leading to improved patient care without compromising data privacy.
Our FL mechanism for glucose prediction follows the FedAvg~\cite{mcmahan2017communication} algorithm, allowing locally trained models to share knowledge about patients’ unique characteristics without transmitting any private data. In an FL training task for glucose prediction, each participating hospital collects relevant data -- glucose levels, insulin intake, meals, physical activity, and sleep patterns -- from their T1D patients via devices like CGM and manually input data. This data is divided into training, validation, and test sets.

Initially, all hospitals and the central server initialize a training model, denoted as $\theta^{\prime}$. The local models at each hospital are aligned with the global model on the central server to ensure a consistent starting point. In the first training round ($r=1$), each hospital uses its own training data to update its local model $\theta_k^{\prime}$. In subsequent rounds ($r > 1$), each hospital updates its local model starting from the previous round's aggregated global model $\hat{\theta}^{r-1}$. After a predefined number of epochs ($E$), the updated local model weights $\theta_k^{r}$ and the number of data samples ($n_k$) are sent to the central server. The central server aggregates the updated model weights from all hospitals using a weighted averaging approach:
\begin{equation*}
\hat{\theta}^{r} = \sum_{k=1}^{K} \beta_k \cdot \theta_k^{r}\text{.}
\label{eq:WeightedAvg}
\end{equation*}
Here, the weight $\beta_k$ is defined as $\frac{n_k}{N}$, where $n_k = |\mathcal{D}_k^{train}|$ is the number of local training data samples for each participant $k$, and $N = \sum_{k=1}^{K} n_k$ is the total number of training data samples across all participants. The aggregated model weights are then broadcast back to the hospitals, allowing them to synchronize their local models with the updated global model. This iterative process continues, with hospitals using the latest model weights for subsequent rounds of local updates. These steps are repeated until the desired number of communication rounds is reached.

\subsection{Blockchain for BCFL}

\subsubsection{Blockchain and Smart Contract}

Blockchain is a decentralized ledger technology that ensures data integrity and transparency across multiple participants in a global peer-to-peer network. Smart contracts are self-executing agreements coded directly onto the blockchain, automating processes upon predefined conditions being satisfied. In BCFL, blockchain and smart contracts play a crucial role by providing a secure and transparent framework for decentralized machine learning. Blockchain can ensure the integrity and traceability of data and model updates, while smart contracts can automate the coordination and validation of contributions from different participants. For instance, smart contracts can be used to implement incentive mechanisms for participants in BCFL.

\subsubsection{Blockchain Tokens}
In our BCFL system, a blockchain token named \FML serves as a quantitative measure of assets. This token can be an indicator of a participant's trustworthiness within our system.  By staking \FML tokens, participants demonstrate their commitment and reliability. Their tokens will be integrated into the incentive mechanism, as shown in Figure~\ref{fig: incentive-mechanism}. Participants risk losing their staked tokens if they act maliciously. Similar to ERC20 tokens, the \FML tokens in our BCFL system are fungible and governed by smart contracts. We assume that all participants have acquired sufficient \FML tokens (e.g., by purchasing \FML from centralized or decentralized exchanges) before joining the BCFL system.



\subsubsection{On-Chain Incentive Mechanism}
As shown in Figure~\ref{fig: incentive-mechanism}, we follow the existing work~\cite{dong2024defending} and adopt an incentive mechanism to reward honest behaviors and penalize malicious behaviors. As demonstrated in~\cite{dong2024defending}, the expected return for malicious participants is negative, and they will ultimately be expelled from the system if they continuously behave maliciously. Different from the existing work~\cite{dong2024defending}, which utilizes smart contracts to perform on-chain aggregation of local model updates, our approach allows voters to execute the aggregation off-chain. This improvement can significantly reduce on-chain costs (e.g., gas fees), particularly when dealing with training models that have a large number of parameters.  

\begin{figure}[t]
\vspace{0.1cm}
\centering
\begin{tcolorbox}[colback=gray!10, colframe=gray!20, width=\columnwidth, sharp corners]
\begin{enumerate}
    \item Participants need to stake a fixed amount (i.e., $\delta$) of \FML tokens into the smart contract to be eligible to participate in the FL training process.

    \item In each round, a participant will be randomly selected to act as either an FL training node or a voter by using a verifiable random function (VRF)~\cite{micali1999verifiable}. 

    \item A training node is responsible for performing local training using its own data and uploading the local model updates to a public database (e.g., IPFS\footnote{https://ipfs.tech/}).

    \item Voters aggregate all the local model updates for that round to obtain the global model update. They evaluate the global model by running it on their evaluation data to propose voting results, i.e., $\mathsf{vote}_i^r \in \{\texttt{support}, \texttt{oppose}\}$.

    \item Based on the majority of the voting result (i.e., $\mathsf{majorVote}^r \in \{\texttt{support}, \texttt{oppose}\}$), participants in that round will either be rewarded or have their FML tokens slashed. Specifically, if a voter's choice, $\mathsf{vote}_i^r$ differs from the majority result (i.e., $\mathsf{vote}_i^t \not= \mathsf{majorVote}^r$), that voter will be penalized; otherwise, they will be rewarded. For all selected training nodes in this round, the majority voting result will determine whether they receive a reward or not.

    \item At the beginning of each round, we check the participants' remaining \FML token balances in the smart contract: if their balance is still larger than a predefined threshold (i.e., $\delta_0$), they will be eligible to continue the following round. Otherwise, they will be removed from the system.

    \item For each round $r$, the finalized global model $\theta^r$ is determined based on the majority voting result. Specifically:

\begin{equation*}
  \theta^r =
    \begin{cases}
      \hat{\theta}^r, & \mathsf{majorVote}^r = \texttt{support} \\
      \theta^{r-1}, & \mathsf{majorVote}^r = \texttt{oppose}
    \end{cases} 
\end{equation*}
\end{enumerate}
\end{tcolorbox}
    \caption{Incentive Mechanism of BCFL.}
    \label{fig: incentive-mechanism}
\end{figure}

Here, $\hat{\theta}^r$ represents the aggregated model proposed by voters in round $r$, and $\theta^{r-1}$ is the last confirmed global model.

\section{Multi-Continental Glucose Prediction Framework Experiments}

\subsection{Experiment Setup}

To expedite the simulations, all experiments were conducted on an NVIDIA RTX4090 GPU. The entire framework, including the complete workflow, was deployed and tested in a decentralized manner on four PCs and one server located in five cities across three continents: Toronto, Canada; Shanghai and Shandong, China; and Newcastle and London, UK. The CPUs used in these machines include Intel Xeon E5-2686, Intel Core i9-13900K, Intel i5-1135, AMD Ryzen 9 7950X, and Apple M3.

In each experiment, five hospitals participate in a federated learning (FL) training task. Each hospital maintains four weeks of glucose-related data for its Type 1 Diabetes (T1D) patients. Data from one week is used for training, with 5\% of this training data set aside for validation. The remaining three weeks of data are used for testing. Overall, we have data from 30 patients, with five patients assigned to each hospital (patient IDs 1 to 25), and five additional patients used exclusively for testing (patient IDs 26 to 30). Specifically, we employed seven methods to train the model: H1Single to H5Single, TotalCentral, and MCGP. H1Single represents single-node training by hospital 1 using its own five patients' data, TotalCentral involves centralized training using data from 25 patients across all five hospitals, and MCGP refers to blockchain-based federated learning where five hospitals collaboratively train a model using data from these 25 patients. We set the number of previous time steps to 24 and the prediction window to 6, representing a total of 30 minutes. Additionally, in the experiment noted with \textit{w/ mal}, we included an extra hospital acting as a malicious participant with data from five fake patients.

We utilize LSTM~\cite{hochreiter1997long} and NNPG~\cite{perez2010artificial} models for our experiments to enable edge computing on local hospital or wearable devices. This light configuration not only enhances performance but also significantly reduces the training load, making the process more efficient and feasible for widespread use in patient care. The learning rate for all experiments is set to 0.000001, using the Adam optimizer with a weight decay of 0.0004. In both centralized and federated training, the local epoch is defined as 5000, making sure the local model can be updated to converge. For FL, we conduct 40 communication rounds to ensure the global model can fully learn the data features from all clients. In each round, 50\% of the total participants are assigned the role of proposer, while the remaining participants are assigned the role of voter. Additionally, we use Chainlink oracles \footnote{https://docs.chain.link/vrf} to provide the Verifiable Random Function (VRF) functionality.

\subsection{Dataset}

The dataset utilized to evaluate the proposed framework consists of \textit{in-silico} data from 30 adult T1D subjects, generated using the UVA/Padova T1D simulator, a tool for glucose level simulation approved by the Food and Drug Administration (FDA)~\cite{man2014uva}. This simulator is a robust and validated method for creating simulated cases. In our study, we employed a modified version of the simulator capable of generating a cohort of T1D cases with configurable meal and insulin information, ensuring sufficient variation among the cases. We created a dataset of 30 unique adult cases, each containing 28 days of data. Each day the data are specifically formatted as a 4-channel matrix, which includes: Glucose (in mg/dL, collected by continuous glucose monitor, 288 samples per day); Meal (carbohydrate in grams, manually input by patients, normally 3-5 samples per day); Insulin (in unit, manually input by patients, normally 3-5 samples per day); Time of the day.
The insulin entries can occur with a meal (administered simultaneously) or without a meal (correction bolus).

In a reward-based system, malicious clients are more likely to earn rewards through simple and crude means, such as generating random false data. To simulate this, we generate fake patients’ data for a malicious hospital aiming to gain rewards with fabricated data. This malicious data is intentionally designed to be significantly outside normal physiological limits, ensuring it is highly unrealistic and disruptive to model training. Specifically, we use the following ranges: glucose values from -10 to 10, insulin values from -5 to 5, measure values from -3 to 3, and time values from -1 to 1. These ranges are selected based on the normal distribution of patient data but are extended far beyond typical values to maximize randomness and extremeness.

\subsection{Evaluation Metrics}
We use root mean square error (RMSE) and mean
absolute relative difference (MARD) for BGLP, as they are the most widely used metrics to measure glucose accuracy. 
\begin{equation}\label{eq:RMSE}
  RMSE = \sqrt{\frac{1}{N}\displaystyle\sum_{k=1}^{N}(g_k-\hat{g}_{k|k-PH})^2},
\end{equation}
where $\hat{g}_{g|k-PH)}$ denotes the prediction results provided by the historical data, and $y$ denotes the reference glucose measurement, and $N$ refers to the data size.
\begin{equation}\label{eq:MARD}
  MARD = \frac{1}{N}\sum_{k=1}^{N}\frac{|\hat{g}_{k|k-PH}-g_k|}{g_k}.
\end{equation}
The RMSE and MARD provide an overall indication of the predictive performance. 

\begin{table*}[ht]
\vspace{0.08in}
\centering
\scalebox{0.85}{
\begin{tabular}{@{}lcccccccccccccccc@{}}
\toprule
\textcolor{tabletitleloss_color}{RMSE Loss} & \multicolumn{7}{c}{Selected Current Patients (in ID)} & \multicolumn{7}{c}{Unseen Patients (in ID)} \\
\cmidrule(lr){2-8} \cmidrule(lr){9-15}
Method & 4 & 7 & 13 & 19 & 23 & Avg & $\Delta_{\text{avg}}$ & 26 & 27 & 28 & 29 & 30 & Avg & $\Delta_{\text{avg}}$ \\
\midrule
H1Single & \textcolor{tab_seen}{9.243} & 12.493 & 11.731 & 9.058 & 12.018 & 10.909 & -1.555 & 11.149 & 14.810 & 9.233 & 9.375 & 10.012 & 10.916 & -1.642 \\
H2Single & 9.874 & \textcolor{tab_seen}{12.026} & 12.052 & 9.231 & 12.243 & 11.085 & -1.704 & 10.872 & 13.609 & 9.544 & 9.466 & 9.970 & 10.692 & -1.418 \\
H3Single & 9.342 & 13.571 & \textcolor{tab_seen}{11.550} & 9.324 & 11.555 & 11.068 & -1.714 & 10.882 & 15.769 & 9.189 & 9.865 & 10.273 & 11.196 & -1.922 \\
H4Single & 10.180 & 12.327 & 11.400 & \textcolor{tab_seen}{9.341} & 11.461 & 10.942 & -1.588 & 10.049 & 13.990 & 9.710 & 9.630 & 10.388 & 10.753 & -1.479 \\
H5Single & 9.674 & 13.013 & 11.501 & 8.946 & \textcolor{tab_seen}{11.540} & 10.935 & -1.581 & 10.419 & 15.212 & 9.592 & 9.450 & 10.469 & 11.028 & -1.754 \\
FedAvg w/ mal & \textcolor{tab_seen}{100.608} & \textcolor{tab_seen}{128.644} & \textcolor{tab_seen}{113.247} & \textcolor{tab_seen}{102.614} & \textcolor{tab_seen}{111.206} & \textcolor{tab_seen}{111.263} & -101.909 & 100.372 & 133.070 & 109.266 & 102.125 & 91.790 & 107.35 & -98.076 \\
TotalCentral w/ mal & \textcolor{tab_seen}{10.080} & \textcolor{tab_seen}{12.555} & \textcolor{tab_seen}{11.836} & \textcolor{tab_seen}{9.210} & \textcolor{tab_seen}{12.087} & \textcolor{tab_seen}{11.154} &  -1.800 & 10.682 & 17.531 & 9.973 & 9.639 & 10.576 & 11.680 & -2.406 \\
TotalCentral & \textcolor{tab_seen}{8.740} & \cellcolor{table_result_2nd} \textcolor{tab_seen}{10.541} & \textcolor{tab_seen}{10.074} & \cellcolor{table_result_2nd} \textcolor{tab_seen}{8.202} & \cellcolor{table_result_2nd} \textcolor{tab_seen}{10.247} & \cellcolor{table_result_2nd} \textcolor{tab_seen}{9.561} & -0.207 & 9.037 & \cellcolor{table_result_best} 12.119 & \cellcolor{table_result_2nd} 8.370 & \cellcolor{table_result_2nd} 8.464 & 9.100 & \cellcolor{table_result_2nd} 9.418 & -0.144 \\
\midrule
MCGP w/ mal & \cellcolor{table_result_best} \textcolor{tab_seen}{8.642} & \textcolor{tab_seen}{10.936} & \cellcolor{table_result_2nd} \textcolor{tab_seen}{10.049} & \textcolor{tab_seen}{8.213} & \textcolor{tab_seen}{10.273} & \textcolor{tab_seen}{9.623} & -0.269 & \cellcolor{table_result_2nd} 8.939 & 13.593 & 8.386 & 8.498 & \cellcolor{table_result_2nd} 8.912 & 9.666 & -0.392 \\
MCGP(ours) & \cellcolor{table_result_2nd} \textcolor{tab_seen}{8.650} & \cellcolor{table_result_best} \textcolor{tab_seen}{10.475} & \cellcolor{table_result_best} \textcolor{tab_seen}{9.707} & \cellcolor{table_result_best} \textcolor{tab_seen}{8.094} & \cellcolor{table_result_best} \textcolor{tab_seen}{9.844} & \cellcolor{table_result_best} \textcolor{tab_seen}{9.354} & -- & \cellcolor{table_result_best} 8.660 & \cellcolor{table_result_2nd} 12.621 & \cellcolor{table_result_best} 8.136 & \cellcolor{table_result_best} 8.224 & \cellcolor{table_result_best} 8.728 & \cellcolor{table_result_best} 9.274 & -- \\
\bottomrule
\toprule
\textcolor{tabletitleloss_color}{MARD Loss} & \multicolumn{7}{c}{Selected Current Patients} & \multicolumn{7}{c}{Unseen Patients} \\
\cmidrule(lr){2-8} \cmidrule(lr){9-15}
Method & 4 & 7 & 13 & 19 & 23 & Avg & $\Delta_{\text{avg}}$ & 26 & 27 & 28 & 29 & 30 & Avg & $\Delta_{\text{avg}}$ \\
\midrule
H1Single & \textcolor{tab_seen}{6.538} & 5.822 & 5.848 & 5.340 & 6.366 & 5.983 & -0.751 & 6.737 & 6.107 & 5.265 & 5.719 & 8.144 & 6.394 & -0.959 \\
H2Single & 6.799 & \textcolor{tab_seen}{5.672} & 5.973 & 5.420 & 6.387 & 6.050 & -0.818 & 6.671 & 5.782 & 5.390 & 5.733 & 7.879 & 6.291 & -0.856 \\
H3Single & 6.574 & 5.865 & \textcolor{tab_seen}{5.898} & 5.477 & 6.342 & 6.031 & -0.799 & 6.671 & 6.138 & 5.234 & 5.979 & 8.232 & 6.451 & -1.016 \\
H4Single & 7.005 & 5.765 & 5.834 & \textcolor{tab_seen}{5.389} & 6.239 & 6.046 & -0.814 & 6.373 & 5.822 & 5.484 & 5.807 & 8.209 & 6.339 & -0.904 \\
H5Single & 6.796 & 5.876 & 5.799 & 5.298 & \textcolor{tab_seen}{6.306} & 6.015 & -0.783 & 6.641 & 6.188 & 5.338 & 5.782 & 8.481 & 6.486 & -1.051 \\
FedAvg w/ mal & \textcolor{tab_seen}{78.861} & \textcolor{tab_seen}{78.050} & \textcolor{tab_seen}{78.008} & \textcolor{tab_seen}{76.768} & \textcolor{tab_seen}{79.100} & \textcolor{tab_seen}{78.157} & -72.925 & 79.364 & 77.981 & 78.502 & 76.903 & 81.648 & 78.880 & -73.445 \\
TotalCentral w/ mal & \textcolor{tab_seen}{7.136} & \textcolor{tab_seen}{5.797} & \textcolor{tab_seen}{6.087} & \textcolor{tab_seen}{5.501} & \textcolor{tab_seen}{6.607} & \textcolor{tab_seen}{6.226} & -0.994 & 6.742 & 6.204 & 5.637 & 6.009 & 8.668 & 6.652 & -1.217 \\
TotalCentral & \textcolor{tab_seen}{6.144} & \cellcolor{table_result_2nd} \textcolor{tab_seen}{5.017} & \textcolor{tab_seen}{5.262} & \textcolor{tab_seen}{4.915} & \textcolor{tab_seen}{5.666} & \textcolor{tab_seen}{5.401} & -0.169 & 5.796 & \cellcolor{table_result_2nd} 5.128 & 4.775 & 5.256 & 7.239 & 5.639 & -0.204 \\
\midrule
MCGP w/ mal & \cellcolor{table_result_2nd} \textcolor{tab_seen}{5.927} & \textcolor{tab_seen}{5.111} & \cellcolor{table_result_2nd} \textcolor{tab_seen}{5.215} & \cellcolor{table_result_2nd} \textcolor{tab_seen}{4.907} & \cellcolor{table_result_2nd} \textcolor{tab_seen}{5.611} & \cellcolor{table_result_2nd} \textcolor{tab_seen}{5.354} & -0.122 & \cellcolor{table_result_2nd} 5.714 & 5.329 & \cellcolor{table_result_2nd} 4.753 & \cellcolor{table_result_2nd} 5.221 & \cellcolor{table_result_2nd} 7.083 & \cellcolor{table_result_2nd} 5.620 & -0.185 \\
MCGP(ours) & \cellcolor{table_result_best} \textcolor{tab_seen}{5.955} & \cellcolor{table_result_best} \textcolor{tab_seen}{4.890} & \cellcolor{table_result_best} \textcolor{tab_seen}{5.081} & \cellcolor{table_result_best} \textcolor{tab_seen}{4.798} & \cellcolor{table_result_best} \textcolor{tab_seen}{5.435} & \cellcolor{table_result_best} \textcolor{tab_seen}{5.232} & -- & \cellcolor{table_result_best} 5.539 & \cellcolor{table_result_best} 5.069 & \cellcolor{table_result_best} 4.602 & \cellcolor{table_result_best} 5.061 & \cellcolor{table_result_best} 6.904 & \cellcolor{table_result_best} 5.435 & -- \\
\bottomrule
\end{tabular}
}

\vspace{0.5mm}
\caption{LSTM model test results on selected (current) and unseen patients’ data. H1Single is trained on patient IDs 1-5, H2Single on IDs 6-10, and so forth. TotalCentral and MCGP are trained on data from patients 1-25, centralized and decentralized, respectively. Evaluation metrics are RMSE (upper table) and MARD (bottom table); lower values indicate better performance. Best results are marked as \colorblockTab[0.3cm]{table_result_best}, second best as \colorblockTab[0.3cm]{table_result_2nd}. Results for current patients are in \textcolor{tab_seen}{purple}.}

\label{tab:LSTM_based}
\vspace{-3mm}
\end{table*}

\begin{figure*}[h]
  \centering
  \includegraphics[width=0.98\linewidth]{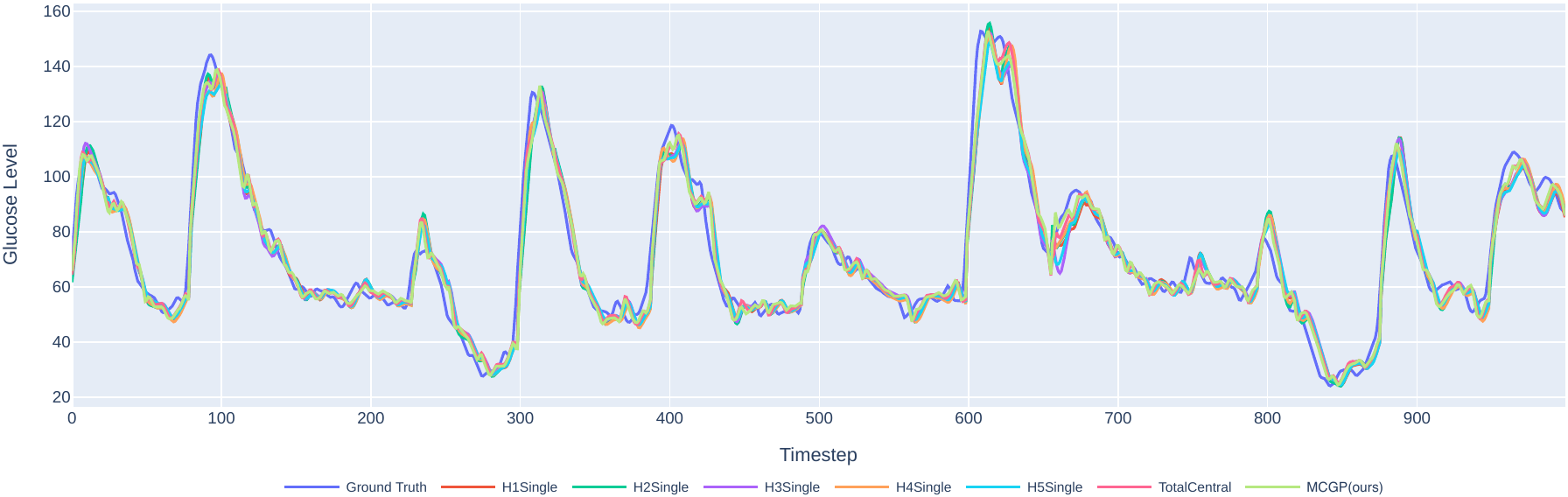}
  \caption{LSTM model test results on Patient23’s local data for the last 1000 points over a three-week period, with glucose levels measured in mg/dL. To better display the values, the glucose levels have been scaled down to half of their original size. The closer to the Ground Truth means the better the result.}
  \label{fig:line_chart_current_patient_23}
\end{figure*}

\subsection{Results}

We produced the results presented in Table~\ref{tab:LSTM_based}, Table~\ref{tab:NNPG_based}, Figure~\ref{fig:line_chart_current_patient_23}, and Figure~\ref{fig:line_chart_unseen_patient_30} by training two different models using various methods on different patients’ data. We validated the effectiveness of our MCGP through cross-validation on known data distributions and demonstrated its generalization and zero-shot capabilities on unseen data.

\subsubsection{\textbf{MCGP Achieves Advanced Model Performance by Breaking Data Silos}}
To demonstrate the performance of MCGP in breaking data silos by integrating the federated learning paradigm, we compared it with standalone single-node training (SNT) conducted by five other hospitals, labeled H1Single to H5Single. The RMSE and MARD losses for selected current patients are reported in Tables~\ref{tab:LSTM_based} and~\ref{tab:NNPG_based}. Our results consistently show that MCGP outperforms the other SNT methods, and even centralised training~(TotalCentral), achieving the lowest RMSE and MARD losses both individually and on average. Besides, MCGP's ability to aggregate global features from all participants enables it to predict more accurately than models trained solely on local data. As illustrated in Figure~\ref{fig:line_chart_current_patient_23}, Participant23’s data, which is part of the H5Single training dataset, does not perform as well as MCGP, particularly at the inflection points of the curve.

\subsubsection{\textbf{MCGP Demonstrates Superior Generalization and Zero-Shot Capabilities}}
In addition to training a model that performs well on all seen patients’ data, we also aim for the model to make accurate predictions on new data. To evaluate the generalization and zero-shot capabilities of MCGP, we focused on its performance with unseen patients (ID 26-30). The results, shown in both tables, highlight MCGP’s ability to generalize well to new, previously unseen data.

\begin{figure*}[h]
  \centering
  \includegraphics[width=0.98\linewidth]{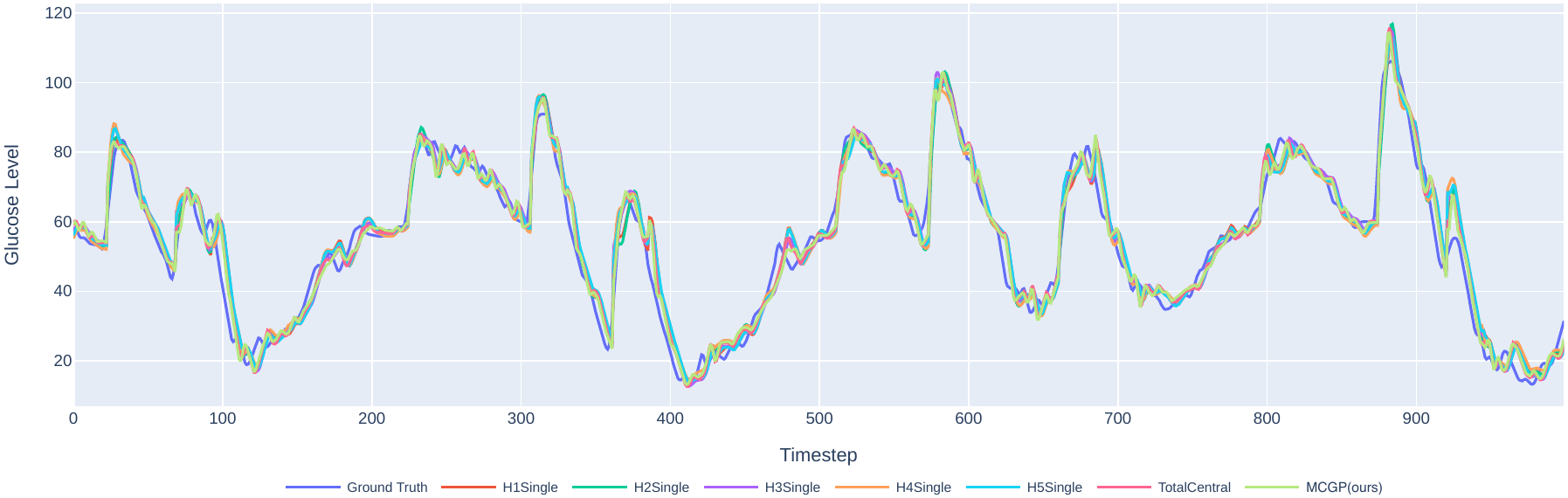}
  \caption{LSTM model test results on unseen data from Patient 30 over the last 1000 points in a three-week period. Close to Ground Truth is better.}
  \label{fig:line_chart_unseen_patient_30}
\end{figure*}

While MCGP can globally learn from all participants’ knowledge, its advantage over TotalCentral diminishes in most tests on unseen patients from current patients, as indicated by $\Delta_{\text{avg}}$ in Table~\ref{tab:LSTM_based} and Table~\ref{tab:NNPG_based}. This suggests that compared to decentralized training, centralized training allows the model to learn detailed characteristics of data from a transparent view. However, since healthcare data is privacy-sensitive and unlikely to be used in centralized training, MCGP maintains significant advantages over other SNT methods, as demonstrated by its more accurate predictions shown in Figure~\ref{fig:line_chart_unseen_patient_30}.


\begin{table*}[ht]
\vspace{0.08in}
\centering
\scalebox{0.88}{
\begin{tabular}{@{}lcccccccccccccc@{}}
\toprule
\textcolor{tabletitleloss_color}{RMSE Loss} & \multicolumn{7}{c}{Selected Current Patients} & \multicolumn{7}{c}{Unseen Patients} \\
\cmidrule(lr){2-8} \cmidrule(lr){9-15}
Method & 4 & 7 & 13 & 19 & 23 & Avg & $\Delta_{\text{avg}}$ & 26 & 27 & 28 & 29 & 30 & Avg & $\Delta_{\text{avg}}$ \\
\midrule
H1Single & \textcolor{tab_seen}{10.480} & 15.837 & 14.359 & 10.432 & 14.540 & 13.130 & -2.862 & 12.437 & 18.672 & 11.108 & 11.083 & 11.822 & 13.024 & -2.861 \\
H2Single & 10.847 & \textcolor{tab_seen}{15.323} & 14.407 & 10.429 & 14.931 & 13.187 & -2.919 & 12.705 & 17.676 & 11.017 & 10.822 & 12.210 & 12.886 & -2.723 \\
H3Single & 10.435 & 16.153 & \textcolor{tab_seen}{14.258} & 10.327 & 14.475 & 13.130 & -2.862 & 12.549 & 19.156 & 10.733 & 10.987 & 11.907 & 13.066 & -2.903 \\
H4Single & 12.033 & 16.235 & 14.205 & \textcolor{tab_seen}{10.446} & 14.835 & 13.551 & -3.283 & 12.350 & 19.148 & 11.195 & 10.874 & 13.306 & 13.375 & -3.212 \\
H5Single & 10.998 & 16.173 & 14.128 & 10.168 & \textcolor{tab_seen}{14.286} & 13.151 & -2.883 & 12.263 & 19.205 & 11.173 & 10.934 & 12.233 & 13.162 & -2.999 \\
FedAvg w/ mal & \textcolor{tab_seen}{52.567} & \textcolor{tab_seen}{60.821} & \textcolor{tab_seen}{55.950} & \textcolor{tab_seen}{50.130} & \textcolor{tab_seen}{54.656} & \textcolor{tab_seen}{54.825} & -44.557 & 51.663 & 65.336 & 52.092 & 50.170 & 47.469 & 53.346 & -43.183 \\
TotalCentral w/ mal & \textcolor{tab_seen}{20.020} & \textcolor{tab_seen}{28.966} & \textcolor{tab_seen}{23.711} & \textcolor{tab_seen}{16.621} & \textcolor{tab_seen}{23.093} & \textcolor{tab_seen}{22.482} & -12.214 & 20.391 & 36.892 & 21.809 & 16.611 & 25.851 & 24.311 & -14.148 \\
TotalCentral & \textcolor{tab_seen}{9.495} & \textcolor{tab_seen}{12.821} & \textcolor{tab_seen}{11.684} & \textcolor{tab_seen}{9.110} & \textcolor{tab_seen}{11.746} & \textcolor{tab_seen}{10.971} & -0.703 & 10.205 & 14.188 & 9.550 & 9.309 & 10.199 & 10.690 & -0.527 \\
\midrule
MCGP w/ mal & \cellcolor{table_result_best} \textcolor{tab_seen}{8.976} & \cellcolor{table_result_best} \textcolor{tab_seen}{11.764} & \cellcolor{table_result_best} \textcolor{tab_seen}{10.685} & \cellcolor{table_result_best} \textcolor{tab_seen}{8.681} & \cellcolor{table_result_best} \textcolor{tab_seen}{10.717} & \cellcolor{table_result_best} \textcolor{tab_seen}{10.165} & 0.103 & \cellcolor{table_result_2nd} 9.787 & \cellcolor{table_result_best} 13.381 & \cellcolor{table_result_best} 8.881 & \cellcolor{table_result_best} 8.872 & \cellcolor{table_result_best} 9.551 & \cellcolor{table_result_best} 10.094 & 0.069 \\
MCGP(ours) & \cellcolor{table_result_2nd} \textcolor{tab_seen}{8.980} & \cellcolor{table_result_2nd} \textcolor{tab_seen}{11.997} & \cellcolor{table_result_2nd} \textcolor{tab_seen}{10.807} & \cellcolor{table_result_2nd} \textcolor{tab_seen}{8.708} & \cellcolor{table_result_2nd} \textcolor{tab_seen}{10.850} & \cellcolor{table_result_2nd} \textcolor{tab_seen}{10.268} & -- & \cellcolor{table_result_best} 9.737 & \cellcolor{table_result_2nd} 13.568 & \cellcolor{table_result_2nd} 8.945 & \cellcolor{table_result_2nd} 8.914 & \cellcolor{table_result_2nd} 9.653 & \cellcolor{table_result_2nd} 10.163 & -- \\
\bottomrule
\toprule
\textcolor{tabletitleloss_color}{MARD Loss} & \multicolumn{7}{c}{Selected Current Patients} & \multicolumn{7}{c}{Unseen Patients} \\
\cmidrule(lr){2-8} \cmidrule(lr){9-15}
Method & 4 & 7 & 13 & 19 & 23 & Avg & $\Delta_{\text{avg}}$ & 26 & 27 & 28 & 29 & 30 & Avg & $\Delta_{\text{avg}}$ \\
\midrule
H1Single & \textcolor{tab_seen}{7.630} & 6.878 & 6.699 & 5.906 & 7.427 & 6.908 & -1.257 & 7.542 & 7.111 & 6.124 & 6.599 & 9.633 & 7.402 & -1.439 \\
H2Single & 7.754 & \textcolor{tab_seen}{6.691} & 6.828 & 6.000 & 7.701 & 6.995 & -1.344 & 7.687 & 6.896 & 6.090 & 6.505 & 9.908 & 7.417 & -1.454 \\
H3Single & 7.624 & 6.914 & \textcolor{tab_seen}{6.775} & 5.934 & 7.501 & 6.950 & -1.299 & 7.617 & 7.143 & 5.999 & 6.678 & 9.697 & 7.427 & -1.464 \\
H4Single & 8.282 & 6.991 & 6.786 & \textcolor{tab_seen}{5.988} & 7.702 & 7.150 & -1.499 & 7.645 & 7.234 & 6.192 & 6.592 & 10.635 & 7.660 & -1.697 \\
H5Single & 7.790 & 6.818 & 6.681 & 5.862 & \textcolor{tab_seen}{7.413} & 6.913 & -1.262 & 7.587 & 7.158 & 6.136 & 6.583 & 10.069 & 7.507 & -1.544 \\
FedAvg w/ mal & \textcolor{tab_seen}{27.586} & \textcolor{tab_seen}{27.989} & \textcolor{tab_seen}{27.696} & \textcolor{tab_seen}{27.258} & \textcolor{tab_seen}{27.807} & \textcolor{tab_seen}{27.667} & -22.016 & 28.860 & 28.198 & 28.688 & 27.691 & 29.552 & 28.600 & -22.637 \\
TotalCentral w/ mal & \textcolor{tab_seen}{14.386} & \textcolor{tab_seen}{12.943} & \textcolor{tab_seen}{11.167} & \textcolor{tab_seen}{9.023} & \textcolor{tab_seen}{12.612} & \textcolor{tab_seen}{12.026} & -6.375 & 13.664 & 14.735 & 12.515 & 10.140 & 23.361 & 14.883 & -8.920 \\
TotalCentral & \textcolor{tab_seen}{6.703} & \textcolor{tab_seen}{5.741} & \textcolor{tab_seen}{5.877} & \textcolor{tab_seen}{5.343} & \textcolor{tab_seen}{6.276} & \textcolor{tab_seen}{5.988} & -0.337 & 6.423 & 5.779 & 5.373 & 5.679 & 8.087 & 6.268 & -0.305 \\
\midrule
MCGP w/ mal & \cellcolor{table_result_best} \textcolor{tab_seen}{6.246} & \cellcolor{table_result_best} \textcolor{tab_seen}{5.304} & \cellcolor{table_result_best} \textcolor{tab_seen}{5.500} & \cellcolor{table_result_best} \textcolor{tab_seen}{5.127} & \cellcolor{table_result_best} \textcolor{tab_seen}{5.853} & \cellcolor{table_result_best} \textcolor{tab_seen}{5.606} & 0.045 & \cellcolor{table_result_best} 6.118 & \cellcolor{table_result_best} 5.556 & \cellcolor{table_result_best} 4.993 & \cellcolor{table_result_best} 5.426 & \cellcolor{table_result_best} 7.555 & \cellcolor{table_result_best} 5.930 & 0.033 \\
MCGP(ours) & \cellcolor{table_result_2nd} \textcolor{tab_seen}{6.283} & \cellcolor{table_result_2nd} \textcolor{tab_seen}{5.389} & \cellcolor{table_result_2nd} \textcolor{tab_seen}{5.539} & \cellcolor{table_result_2nd} \textcolor{tab_seen}{5.144} & \cellcolor{table_result_2nd} \textcolor{tab_seen}{5.899} & \cellcolor{table_result_2nd} \textcolor{tab_seen}{5.651} & -- & \cellcolor{table_result_2nd} 6.126 & \cellcolor{table_result_2nd} 5.601 & \cellcolor{table_result_2nd} 5.045 & \cellcolor{table_result_2nd} 5.459 & \cellcolor{table_result_2nd} 7.583 & \cellcolor{table_result_2nd} 5.963 & -- \\
\bottomrule
\end{tabular}
}
\vspace{0.5mm}
\caption{Test results of NNPG model.}
\label{tab:NNPG_based}
\vspace{-3mm}
\end{table*}

\subsubsection{\textbf{MCGP can resist the impact of malicious participant}}
In a real decentralized Federated Learning (FL) scenario, maintaining the performance of the global model is crucial, especially for tasks sensitive to prediction accuracy, such as blood glucose prediction. Our experiments, which include malicious participants (indicated as \textit{w/ mal}) and are summarized in Table~\ref{tab:LSTM_based} and Table~\ref{tab:NNPG_based}, demonstrate that models trained using our MCGP are largely unaffected by these malicious participants, maintaining performance comparable to scenarios without malicious participants. In contrast, the FedAvg method is significantly impacted by malicious clients, for example, with RMSE loss deteriorating to 111.263 compared to 9.354 for MCGP, and MARD loss worsening to 78.157 compared to 5.232 for MCGP, as illustrated in Tables~\ref{tab:LSTM_based}. Additionally, in centralized training, where data from malicious clients constitutes 1/6 of the total, the centralized approach (TotalCentral w/ mal) shows some resilience but still performs worse compared to pure centralized training (TotalCentral), particularly for the NNPG model, as shown in Table~\ref{tab:NNPG_based}.
\subsection{Discussion}
\subsubsection{\textbf{Beyond Expected Results}}
Federated Learning has demonstrated its efficiency in computer vision tasks~\cite{ye2023feddisco,sun2024rehearsal}, but it continues to face challenges with distributed data optimization. Typically, models trained using FL achieve performance close to that of models trained on centralized data but rarely surpass them due to the optimization issues arising from data distribution silos~\cite{zhao2018federated}. However, in our research, we focus on time-series data, which is lower-dimensional compared to image data. As a result, the models are powerful enough to learn features effectively.

Our study yielded surprisingly positive results, with the FL-trained models outperforming those trained on centralized data. This can be attributed to two main factors. First, the inherent simplicity of time-series data allows models to learn relevant features efficiently, making the differences between FL and centralized training negligible in this context. Second, more importantly, the glucose prediction task we tackled involves data with distinct personal characteristics. In centralized training, aggregating all individual features can obscure personal nuances, making it difficult for the model to differentiate between individual characteristics. Conversely, FL optimizes the model within smaller, localized datasets, allowing it to more clearly capture and distinguish personal features.

This personalized approach in FL not only reduces overfitting to generalized patterns but also enhances model’s ability to generalize across different individuals. Consequently, FL-trained models exhibit superior performance in tasks requiring attention to individual variations, such as glucose prediction, compared to their centrally trained counterparts. This highlights the potential of FL in applications where personalized data characteristics are crucial for accurate predictions.

\subsubsection{\textbf{Decentralised System Orchestration Barriers}}
During the real deployment testing stage of our MCGP across five locations, we encountered a networking issue due to system environment heterogeneity among devices. Specifically, two participants ran our MCGP on different platforms: one used a VMWare Workstation\footnote{https://www.vmware.com/content/vmware/vmware-published-sites/us/products/desktop-hypervisor.html.html.html} virtual machine~(VM), while the other utilized Microsoft Windows native WSL2\footnote{https://learn.microsoft.com/en-us/windows/wsl/about} VM.

Specifically, for the VMware VM, the network is virtualized and isolated from the local system, necessitating meticulous network configuration before running the MCGP. This added an extra layer of complexity, as users needed to ensure that all network settings were correctly configured, which required significant technical knowledge and effort. For the WSL2 VM, the participant was located in Mainland China, where the Great 
Firewall\footnote{https://en.wikipedia.org/wiki/Great\_Firewall} restricts access to the resources required by the MCGP. Consequently, the user needed to use a VPN to access these sites. However, configuring a VPN within the WSL2 VM proved to be particularly challenging due to compatibility issues and the additional overhead of managing VPN connections in this environment.

These challenges highlight common barriers in deploying an automated decentralized system orchestration. From a real engineering perspective, designing an adaptive mechanism to address such issues is crucial for ensuring smooth operation across diverse environments and enhancing the robustness of decentralized systems. 


\section{Conclusion and Future Work}
In this work, we proposed a Multi-Continental Glucose Prediction (MCGP) framework using blockchain-enabled federated learning to address the challenges of data privacy and sharing in healthcare. Our experimental results demonstrated that the MCGP framework effectively preserves data privacy while significantly improving prediction accuracy compared to traditional methods. This approach facilitates global collaboration, allowing healthcare institutions to contribute to model training without sharing sensitive healthcare data directly. Future work will focus on extending the framework to other chronic diseases, enhancing privacy with differential privacy techniques, and improving computational efficiency. 

\bibliographystyle{ieeetr}
\bibliography{references}

\end{document}